\documentclass[conference]{IEEEtran}
\IEEEoverridecommandlockouts

\usepackage[utf8]{inputenc}   
\usepackage{graphicx}
\usepackage{amsmath,amssymb,amsfonts}
\usepackage{algorithm}
\usepackage{algorithmic}
\usepackage{booktabs}
\usepackage{xcolor}
\usepackage{textcomp}
\usepackage{cite}
\usepackage[most]{tcolorbox}
\usepackage{url}

\begin{document}

\title{Natural Language Instructions for Scene-Responsive Human-in-the-Loop Motion Planning in Autonomous Driving using Vision-Language-Action Models}

\author{\IEEEauthorblockN{Angel Martinez-Sanchez}
\IEEEauthorblockA{\textit{University of California, Merced} \\}
\and
\IEEEauthorblockN{Parthib Roy}
\IEEEauthorblockA{\textit{University of California, Merced} \\}
\and
\IEEEauthorblockN{Ross Greer}
\IEEEauthorblockA{\textit{University of California, Merced} \\}
}

\maketitle

\begin{abstract}
Instruction-grounded driving, where passenger language guides trajectory planning, requires vehicles to understand intent before motion. However, most prior instruction-following planners rely on simulation or fixed command vocabularies, limiting real-world generalization. doScenes, the first real-world dataset linking free-form instructions (with referentiality) to nuScenes ground-truth motion, enables instruction-conditioned planning. In this work, we adapt OpenEMMA, an open-source MLLM-based end-to-end driving framework that ingests front-camera views and ego-state and outputs 10-step speed–curvature trajectories, to this setting, presenting a reproducible instruction-conditioned baseline on doScenes and investigate the effects of human instruction prompts on predicted driving behavior. We integrate doScenes directives as passenger-style prompts within OpenEMMA’s vision–language interface, enabling linguistic conditioning before trajectory generation. 
Evaluated on 849 annotated scenes using ADE, we observe that instruction conditioning substantially improves robustness by preventing extreme baseline failures, yielding a 98.7\% reduction in mean ADE. When such outliers are removed, instructions still influence trajectory alignment, with well-phrased prompts improving ADE by up to 5.1\%. 
We use this analysis to discuss what makes a ``good'' instruction for the OpenEMMA framework. We release the evaluation prompts and scripts to establish a reproducible baseline for instruction-aware planning. GitHub: \url{https://github.com/Mi3-Lab/doScenes-VLM-Planning}
\end{abstract}

\begin{IEEEkeywords}
safe autonomous driving, human-robot interaction, vision language action models, motion planning
\end{IEEEkeywords}


\section{Introduction}               

In the emerging era of autonomous driving, vehicles must operate in dynamic and unpredictable environments \cite{wang2024drivedreamer}. Beyond perception and planning (e.g., understanding the environment and deciding how to move within it), for safe open-world responsiveness, autonomous vehicles (AVs) must also interpret and act on natural-language instructions that change goals and constraints in real time \cite{greer2024towards, roh2020conditional}. In practice, passenger instructions can directly reshape the motion plan—for example, “drop me just past the intersection, behind the gray car, in the open curb space.” Executing such a directive not only requires perception of the surrounding traffic but also referent grounding (``gray car''), temporal anchoring (``past the intersection''), constraint enforcement (legal curb/bike-lane rules), and an update of the planner’s objective. While research has emphasized perception, mapping, and prediction (e.g., detection \cite{greer2025language} and trajectory forecasting \cite{keskar2025mtr}), comparatively little attention has been paid to how free-form, imperative language directly influences motion behavior, which is why bridging this gap between language understanding and continuous control is essential for developing AVs that can follow real-time verbal input from passengers.

\begin{figure}
    \centering
    \includegraphics[width=0.85\linewidth]{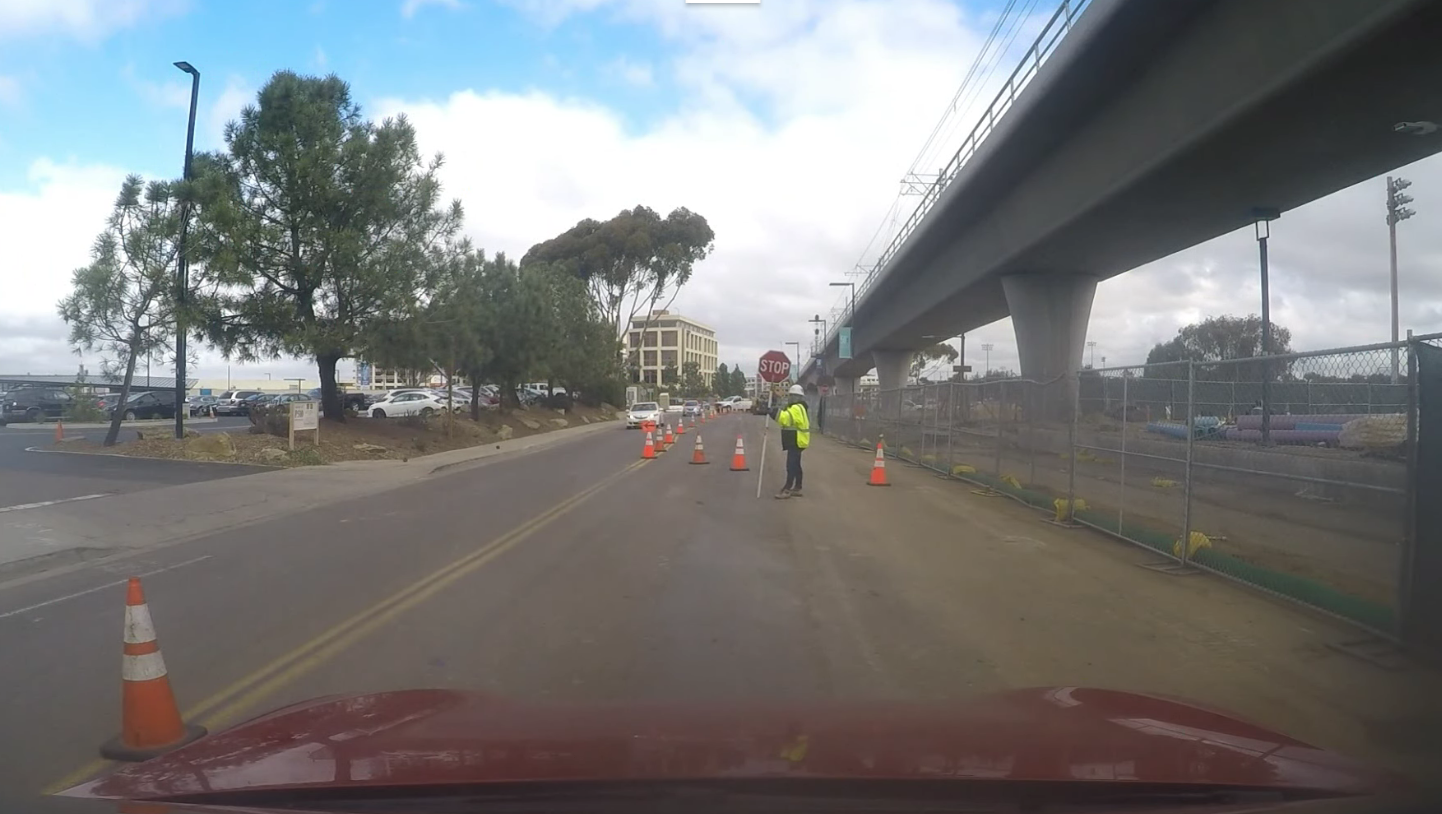}
    \caption{Dynamic or suddenly-unstructured driving scenes create situations which call upon a human-in-the-loop to offer an instruction to the autonomous vehicle for appropriate scene reasoning and planning. This is especially relevant in ambiguous scenes, such as the scene depicted above, where the vehicle could reasonably ``turn left into the parking lot'' or ``proceed straight but left of the cones'' depending on what the passenger determines to be safe and goal-aligned. We present a method for integrating natural language instructions into a VLM-driven motion planner, and demonstrate the effectiveness of instruction-informed trajectories in emulating ground truth.}
    \label{fig:placeholder}
\end{figure}


In response to this gap, the doScenes dataset \cite{doScenes} was released in January 2025. doScenes was the first public dataset pairing real-world multimodal sensor data from driving scenes, trajectory ground truth annotations, and associated natural-language passenger instructions (e.g., “Turn left then turn slightly right to avoid the road obstruction”). A natural research question following the dataset release was the utility of this new form of data, which combines a natural language instruction and a representation of a driving environment and a planned motion. While current vision-language-action (VLA) models in autonomous driving focus on using visual question-answering (VQA) or end-to-end methods to output trajectories towards a determined goal, waypoint, or high-level directive (e.g., ``turn left'', ``turn right'', ``go straight'') \cite{park2024vlaad, arai2025covla, gopalkrishnanmulti,renz2025simlingo, keskar2025evaluating, shriram2025towards}, the challenge of doScenes is novel in its free-form annotation style, where annotations may be grounded to scene elements, exist on sequential timescales, and contain fine-grained details of expected driving patterns, as illustrated in Figure \ref{fig:placeholder}. Thus, in this paper, we present the first analysis of fine-grained instruction-conditioned planning for autonomous driving by adapting OpenEMMA \cite{openEMMA}, an open-source, end-to-end framework built on multimodal large language models (MLLMs) that generates driving trajectories from visual inputs through Chain-of-Thought (CoT) reasoning. Specifically, we augment OpenEMMA with passenger-style natural language instructions from the doScenes dataset and introduce language as an additional input, alongside its existing front-camera stream and ego-vehicle history. From these inputs, the model reasons, predicts future speed–curvature sequences, and integrates them into a driving trajectory. We then  compare against an instruction-agnostic baseline using average displacement error (ADE) to answer a focused question: Does natural language conditioning meaningfully steer the plan? 



Our contributions are:
\begin{itemize}
  \item \textbf{First application of fine-grained instruction conditioning for AV planning}: We utilize passenger-style, free-form doScenes instructions to steer nuScenes trajectories produced by OpenEMMA and evaluate the effect with ADE.
  \item \textbf{Baseline, Instruction-level, and failure-mode analysis:} In addition to comparing the effects of instructed planning versus generic, non-prompted driving baselines, we also analyze patterns in human instructions such as length, clarity, grounding, and referentiality, summarizing their empirical effects on predicted trajectories. We further identify scene-level failure modes such as out-of-bounds predictions and demonstrate cases where well-phrased instructions correct these behavior.
  \item \textbf{Reproducible baseline}: We release code, prompts, and evaluation scripts so others can reproduce the setup and analyze additional instruction-style prompts.

\end{itemize}

This work marks the first application of doScenes to the autonomous driving planning task, demonstrating that natural-language instructions can directly shape a vehicle’s motion plan. This serves as an important proof of practicality for doScenes and establishes it as an active benchmark for instruction-conditioned control rather than a static training dataset. 

\section{Related Research}

\subsection{Autonomous Driving Datasets and Natural Language}

nuScenes \cite{nuScenes} is the first AV dataset to feature a complete 360 degree Field of View (FOV) sensor suite, released in 2019. It provides 1,000 real-world driving scenes collected from Boston and Singapore, two highly challenging and dense traffic environments. In addition to typical weather conditions, the scenes also feature a mix of nighttime and rainy condition data, and are captured from two Renault Zoe supermini AV vehicles, equipped with 6 cameras, 5 radars, and 1 LiDAR. Annotations are done at 2 Hz, resulting in 1.4 million 3D bounding boxes across 23 object classes. This dataset is utilized by both the doScenes dataset and the OpenEMMA framework, thus serves as a critical resource for this work.


The doScenes dataset \cite{doScenes} was designed to provide data for systems to learn a relation between instructions and vehicle motion. The dataset is applied on the nuScenes dataset and focuses on short-term interactions (on the order of $ \le 10$ seconds of motion).
For each of the nuScenes clips, the ``taxi test'' heuristic is applied: ``If you were in a taxi, what instruction would you give to initiate this motion?'' The dataset contains 3,924 annotations, with the following distributions: 64.2\% Non-Referential, 11.1\% Dynamic, 23.5\% Static, 6.8\% with both Dynamic \& Static. The annotations were produced by six independent annotators. 


\subsection{Language-Conditioned Planning Models}
DriveMLM \cite{driveMLM} is a multimodal driving framework, published in 2023, that combines visual perception with language reasoning in the CARLA \cite{carla} simulator. Its language interface is designed around high-level behavioral commands, such as “change lane,” “keep lane,” or “slow down,” which are mapped into discrete planning states used by modular stacks like Apollo \cite{apollo}. Because the model predicts behavior classes rather than continuous trajectories, DriveMLM only accepts short, templated commands and cannot process the kind of free-form, referential, multi-step instructions found in doScenes (e.g., “merge right after the silver van passes, then stop before the crosswalk”). This difference is fundamental, as the DriveMLM’s architecture expects categorical intent, not natural-language directives that directly reshape 10-step speed–curvature motion. As a result, DriveMLM is not ready for doScenes-style instructions. As it entirely runs in simulation-form and relies on a fixed set of predefined commands, there is not yet an ability to connect free-form language to real-world vehicle motion. This gap highlights the difference in our work: we are the first to apply passenger-style instructions to a real-world trajectory prediction model.

Similarly, GPT-Driver \cite{gptDriver} reformulates motion planning as a language modeling problem. Rather than consuming passenger-issued instructions as input, it converts perception outputs--object positions, velocities, and the ego vehicle’s states--into textual tokens. Thereafter, a large language model is asked to generate the next set of tokens in the sequence as future waypoints. For example, a scene might be encoded as “car at (-8.5, 0.1), velocity (0.0, 1.4)”--to which the model then predicts the upcoming coordinates in the same textual format. This makes the forecasting task feel similar to next-word prediction, but the system is never conditioned on what a passenger actually says; it reasons over a generated scene summary, not human instruction. As a result, GPT-Driver is not designed to follow doScenes-style, free-form directives or adapt its plan based on what a passenger says.

LMDrive \cite{shao2024lmdrive}, another language-guided framework, unifies the two prior ideas: it uses an LLM to generate waypoints, as in GPT-Driver, and it executed them in a closed-loop environment within CARLA, as in DriveLML. The LLM-generated waypoints are immediately tracked by CARLA's PID controllers, so each LLM output is converted into steering, throttle, and brake commands that update the environment and determine the next observation. With this closed-loop setup, the task is moved closer to real-world driving, where the agent must remain stable over long horizons, recover from its own mistakes, and handle the cumulative effect of its language-conditioned planning, rather than simply scoring well on a single open loop trajectory. Similar to DriveMLM, LMDrive’s simulation-based environment prevents the direct use of the doScenes dataset, which is tied to real nuScenes sensor logs. However, this does not prevent incorporating doScenes-style natural language; such instructions could be used if a CARLA dataset were constructed with free-form, passenger-level annotations, making this a natural direction for future work.


OpenEMMA is an open-source implementation of Waymo's EMMA \cite{waymoEMMA}, and leverages Vision Language Models (VLMs) to integrate text and front-view camera inputs. In the original implementation, OpenEMMA was integrated and validated with the nuScenes dataset. Additionally, OpenEMMA processes each nuScenes scene through three reasoning stages before producing a trajectory video.
\begin{enumerate}
    \item \textbf{Scene description}: Collects images (front view camera frames) and retrieves a summarized, human-readable scene description from the VLM Description prompts the VLM to focus on traffic lights, other vehicles or pedestrians, and lane markings. At the end, it calls VLM-inference to fetch the final textual output. Simply, Scene Description asks the model to provide a general ``what is going on in the scene?'' summary and focuses on traffic lights, other cars, and lane markings.
    \item \textbf{Object identification}: Like scene description, collects images, but also retrieves a short textual listing of important road users from the VLM. It prompts the VLM to list two or three of them specifying their location and explaining why they matter.
    \item \textbf{Intent Estimation} - Determines and updates the ego car's driving intent from the VLM. If there is no previous intent the prompt focuses on whether the car will turn left, turn right, or go straight, and at what speed. If there is a previous intent, it checks if the intent remains valid based on new observation.
\end{enumerate}

After the completion of all three stages, the model predicts a 10-step ego-vehicle trajectory in x,y coordinates, suitable for open-loop evaluation with nuScenes trajectory data.

Importantly, OpenEMMA sees, recognizes, and reasons over the scene; however, it lacks the ability to listen. In this paper, we add that missing sense: for the first time, the model is guided not only by what is in front of the car, but also by what the passenger instructs.

\section{Methodology}


This work investigates how motion planning by AV systems can be altered with natural-language, specifically doScenes, instructions. We integrate doScenes into OpenEMMA, an open-source end-to-end driving framework that predicts future speed and curvature of the ego vehicle, which are then integrated into the trajectory generation process.

\subsection{Data Source}
As doScenes was developed as an extension of the nuScenes \cite{nuScenes} dataset, we operate entirely within nuScenes and use the front-camera images that are expected by OpenEMMA. At the time of inference, we augment each forward pass with the corresponding doScenes instruction, which allows us to evaluate the effect of language without extensively modifying the underlying OpenEMMA architecture.

\subsection{Instruction Integration into OpenEMMA}
OpenEMMA processes video frames, ego-vehicle states, and map layers through a vision–language interface before generating a 10-step curvature and velocity plan. To study the effect of natural-language conditioning, we inject \textit{doScenes} instructions as passenger directives within the model’s prompt. Specifically, each instruction is inserted into the existing scene-description template as follows:


\begin{tcolorbox}[colback=gray!10, colframe=gray!50, boxrule=0.3pt, arc=2pt, left=6pt, right=6pt, top=4pt, bottom=4pt]
\texttt{The passenger says: "<doScenes\_instruction>". Always prioritize the passenger’s instruction unless it is unsafe; if complying is unsafe, briefly explain and choose the safest alternative.}
\end{tcolorbox}

Treating the instruction as the single controlled change in the original OpenEMMA pipeline keeps the experiment interpretable by ensuring that any output variation can be traced directly to the injected language. We acknowledge that, on any given scene, the model may vary slightly due to VLM stochasticity \cite{peeperkorn2024temperature, li2025exploring}, but over the full set of scenes, consistent shifts in the output would indicate that the change comes from the instruction rather than random noise.


\subsection{Application: Instruction-Conditioned Trajectory Generation}
To evaluate the effect of natural-language instructions on vehicle motion, we compare OpenEMMA’s predicted trajectories under two conditions: (1) without instruction input and (2) with \textit{doScenes}-derived passenger directives. To ensure that the only difference between runs is the inclusion of natural-language instructions, both conditions are evaluated on the same nuScenes scenes. After a nuScenes scene has been run under both conditions, OpenEMMA outputs qualitative logs (text file with scene, object, and intent descriptions), as well as quantitative motion data (predicted curvature-velocity pairs) that together form a 10-step trajectory.

As a note, many nuScenes clips contain multiple doScenes annotations, since different annotators provided similar but unique instructions for the same scene. We treat each instruction as an independent evaluation, meaning the same nuScenes clip is passed through OpenEMMA multiple times, once per instruction. All ADE scores are computed at the instruction level, not the scene level.

\paragraph{Evaluation Metric}
We measure trajectory alignment by using \textit{Average Displacement Error (ADE)}, which measures geometric similarity between predicted and ground-truth trajectories, and is the standard metric in nuScenes trajectory evaluation \cite{nuscenes_traj, greer2021trajectory}. 

\[
\text{ADE} = \frac{1}{T} \sum_{t=1}^{T} \lVert \hat{y}_t - y_t \rVert_2
\]
where $T$ is the prediction horizon, $\hat{y}_t$ is the predicted position at time $t$, and $y_t$ is the ground-truth position.

ADE provides a direct and simple way to detect whether the doScenes prompts have any effect on the trajectory output. Even a small difference between the two conditions would indicate that the doScenes instruction influences the predicted trajectory at a basic level. We also produce trajectory visualizations for both settings to qualitatively compare how the predicted ego-vehicle path changes.


\subsection{Scene Selection}
The nuScenes dataset contains roughly 1,000 real-world driving clips. Of all doScenes annotations (3,924), 1,423 (36\%) include instructions with clear language–trajectory alignment (e.g., “Turn right at the second stoplight”). The remaining clips contain non-actionable instructions (e.g., waiting at a red light or driving straight with no intervention), and annotators were instructed to provide no instruction in cases where a driver continuing the present behavior would already be expected. As the non-actionable instructions cannot induce a change to the motion plan's output, we restrict the evaluation to only the actionable subset. This way, we are able to ensure that every evaluated sample contains sensor data paired with a valid passenger-style instruction.

\subsection{Multimodal Large Language Model Selection}
We use LLaVA-1.6-Mistral-7B \cite{llava1.6}—chosen for its practicality and reproducibility. Firstly, OpenEMMA natively builds around LLaVA: its scene-description, object-identification, and intent-estimation prompts rely on LLaVA’s multimodal chat template, image-token format, and tokenizer integration. Therefore, using LLaVA allows us to inject doScenes instructions directly into the reasoning pipeline without architecture modifications. Secondly, LLaVA is fully open-source and runs locally, which removes dependence on proprietary APIs and ensures that our setup can be easily reproduced by others. 

As a note, for this baseline study, we intentionally evaluate only with this model rather than stronger VLMs (e.g., Qwen2.5-VL \cite{qwen2.5}), which we discuss later as future work.

\subsection{Hardware}
The processing of nuScenes clips were split across two compute environments: one equipped with a NVIDIA RTX 4090 GPU and another with an NVIDIA RTX 6000 Ada GPU. For both machines, parallelization was not possible due to the LLaVA model fully occupying the GPU during inference, and this VLM inference step was the dominant computational bottleneck. Despite this real-time limitation, parallel research in model optimization and efficiency suggest that latency will continue to improve for foundation models on edge devices \cite{he2025road, xu2025resource}. Overall, each clip required, on average, ~20 minutes to process, and the full evaluation ran over 7 days.


\section{Results \& Discussion}

To evaluate the influence of natural-language conditioning on trajectory prediction, we computed the ADE for each scene without language input (\textit{No Instr.}) and with \textit{doScenes} instructions. Since each scene may include multiple annotations by different annotators, we determined the lowest (\textit{Best ADE}), mean (\textit{Avg ADE}), and highest (\textit{Worst ADE}) ADE across all annotations for the same scene. These values can reveal how prompt phrasing can affect model's performance.
When evaluated across all 849 annotated scenes, the no-instruction baseline produces several extreme outliers that heavily skewed the mean, resulting in a 98.7\% improvement in ADE for the instruction-conditioned planner. After applying a 97.5\textsuperscript{th}-percentile (Q97.5) filter to remove roughly twenty high-error outlier scenes, the overall improvement dropped to 5.1\%, revealing clearer patterns linked to prompt quality as shown in Table~\ref{tab:ade_summary_grouped_both}

\begin{table}
\caption{ADE comparison between baseline condition and instruction-conditions runs with all and filtered scenes}
\centering
\resizebox{\columnwidth}{!}{
\begin{tabular}{lccccc}
\toprule
& \multicolumn{1}{c}{\textbf{No Instr.}} & \multicolumn{3}{c}{\textbf{doScenes}} \\
\cmidrule(lr){2-2}\cmidrule(lr){3-5}
\textbf{} & \textbf{Avg ADE} & \textbf{Best ADE} & \textbf{Avg ADE} & \textbf{Worst ADE} \\
\midrule
\textbf{Mean (All)} & 6201.443 & \textbf{9.999} & 78.527 & 151.420 \\
\textbf{Mean (Q97.5)} & 2.879 & \textbf{2.732} & 2.929 & 3.11 \\
\bottomrule
\end{tabular}
}
\label{tab:ade_summary_grouped_both}
\end{table}

Across all scenes, instruction-based annotations yield better overall results than no instruction. Yet if we trim the results to filter out any outliers, only the best ADE score from \textit{doScenes}  preforms better than the no instruction baseline. These two key observations highlight: (1) the VLM model occasionally produces unrealistic predictions causing outlier ADE scores, and (2) prompt phrasing has an impact on the results of the predicted trajectory.

\subsection{Outlier Scenes}

\begin{figure}
    \centering
    \includegraphics[width=0.95\columnwidth]{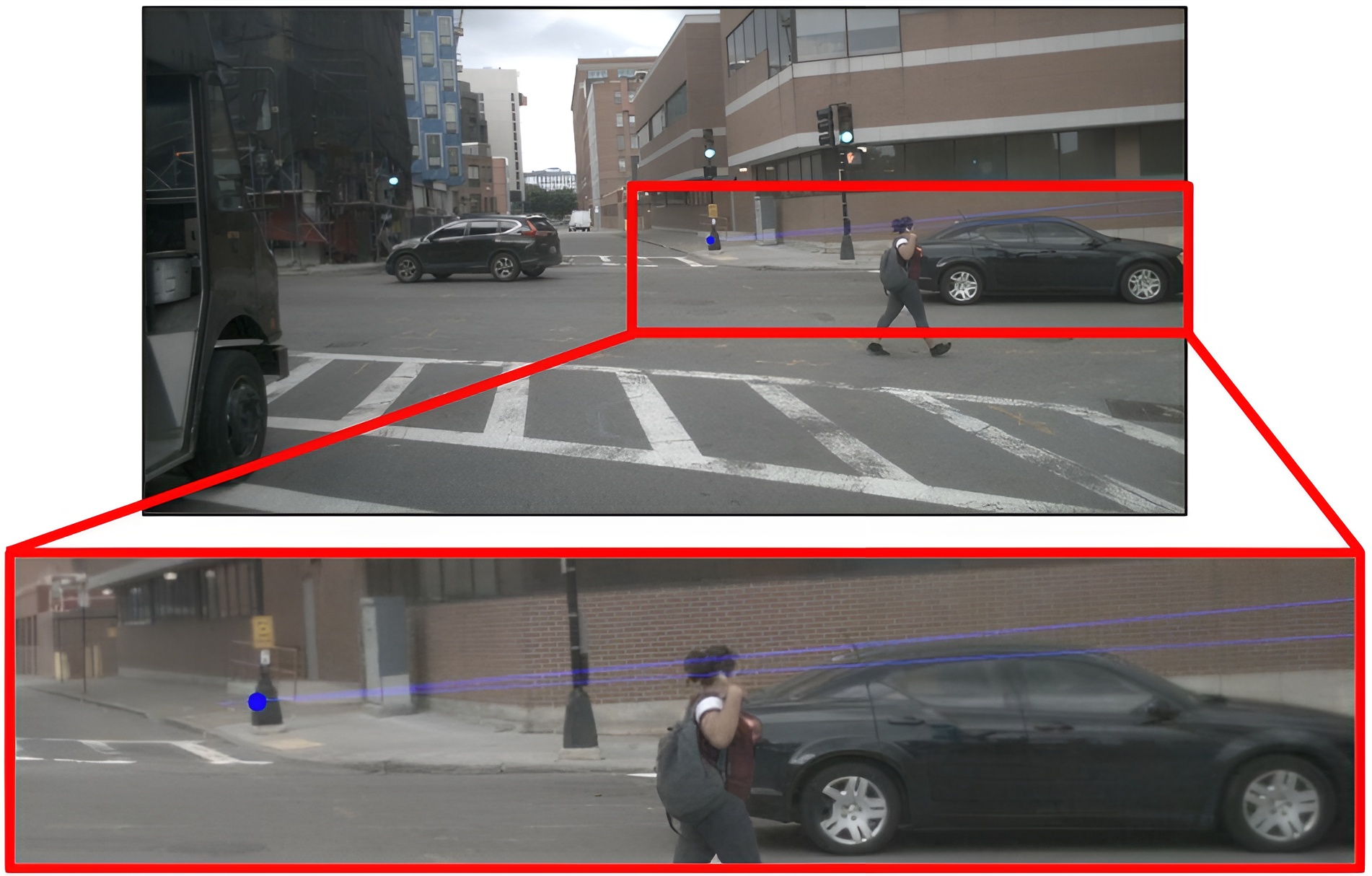}
    \caption{Example of the OpenEMMA baseline model making predictions of points which fall outside the scene boundary.}
    \label{fig:scene760_woinstr}
\end{figure}

When investigating the scenes that caused abnormal ADE scores, it was determined that the VLM model would occasionally predict  waypoints outside the captured scene environment. Figure~\ref{fig:scene760_woinstr} illustrates an example where a predicted waypoint is set across the intersection and is connecting to a trajectory line extending outside the scene boundary. This behavior was more frequent when no instructions were provided. As shown in Table~\ref{tab:ade_summary_grouped_both}, any form of doScenes prompt instruction significantly improved performance when the outlier scenes were included.

These outlier cases often correlate to scenes where the ego vehicle was expected to remain stationary for an extended period. The doScenes instructions helped the model understand that it should remain stationary until the required conditions were met (e.g., a red light turning green or a pedestrian/vehicle moving out of the ego car's way). The same scene from Figure~\ref{fig:scene760_woinstr} did not experience outlying waypoints when waiting before the intersection using the doScenes annotations.

\subsection{Prompt Phrasing}

With the results from Table~\ref{tab:ade_summary_grouped_both}, it was apparent that subtle differences in the instruction can cause better or worse results compared to running OpenEMMA without any instruction input. 
Given this result, we analyzed patterns in the instructions themselves to better understand what aspects of language contribute to predicting more accurate trajectories. 

Table~\ref{tab:wordlength_ade} summarizes the relationship between instruction length and prediction accuracy. Although the longest prompts (19+ words) achieves the lowest overall ADE, a Typical (9-12 words) yields the greatest improvement relative to the same scene without instructions. This finding suggests that shorter instructions leads to a degrade in performance, while too long if an instruction can be too verbose for a VLM to produce the most improved scores. 


\begin{table}
\caption{Average ADE grouped by instruction length buckets. We report OpenEMMA performance without language conditioning and with doScenes instructions.}
\centering
\resizebox{\columnwidth}{!}{
\begin{tabular}{lccc}
\toprule
\textbf{Bucket} & \textbf{Word Range} & \textbf{ADE (No Instr. Q97.5)} & \textbf{ADE (doScenes Q97.5)} \\
\midrule
Ultra-Short & 0–4   & 3.001 & 3.323\\
Short       & 5–8   & 3.002 & 3.076 \\
Typical     & 9–12  & 2.916 & 2.887\\
Descriptive & 13–18 & 2.925 & 2.902\\
Long        & 19+   & 2.795 & 2.784\\
\bottomrule
\end{tabular}
}
\vspace{5pt}
\label{tab:wordlength_ade}
\end{table}

Building on this observation, we next examine whether the type of referentiality used in the instruction affects the trajectory accuracy. Each doScenes annotation can direct the ego vehicle using language that references static elements (e.g., road markings, intersections, or signs), dynamic elements (e.g., vehicles or pedestrians), or a combination of both.  Table~\ref{tab:referentiality_vs_ade} reports the average ADE for each referentially category. 
Instructions that reference dynamic objects resulted in the lowest overall ADE values. This suggested that OpenEMMA benefits from prompt cues tied to moving objects that provide temporal or relational context. Examples of effective dynamic instructions include \textit{“Follow the yellow car”} or \textit{“Slow down for the pedestrians”}. In contrast, non-referential or purely static prompts (e.g., \textit{“Go straight”}, \textit{“Turn left at the intersection”}) tend to produce higher ADEs given that they provide limited information about how the ego vehicle should respond to surrounding objects.


\begin{table}
\centering
\caption{ADE comparison across referentiality categories.}
\resizebox{\columnwidth}{!}{
\begin{tabular}{lccc}
\toprule
\textbf{Referentiality} & \textbf{ADE (No Instr. Q97.5)} & \textbf{ADE (doScenes Q97.5)} \\
\midrule
None (Non-ref)      & 3.014 & 3.397 \\
Static Only         & 3.054 & 3.027 \\
Dynamic Only        & 2.830 & \textbf{2.764}\\
Static + Dynamic    & 2.829 & 2.783 \\
\bottomrule
\end{tabular}
}
\label{tab:referentiality_vs_ade}
\end{table}

\begin{figure*}
    \centering
    \begin{tabular}{cc}
        \toprule
        \textbf{Scene 238} & \textbf{Scene 642} \\
        \midrule
        \includegraphics[width=0.425\textwidth]{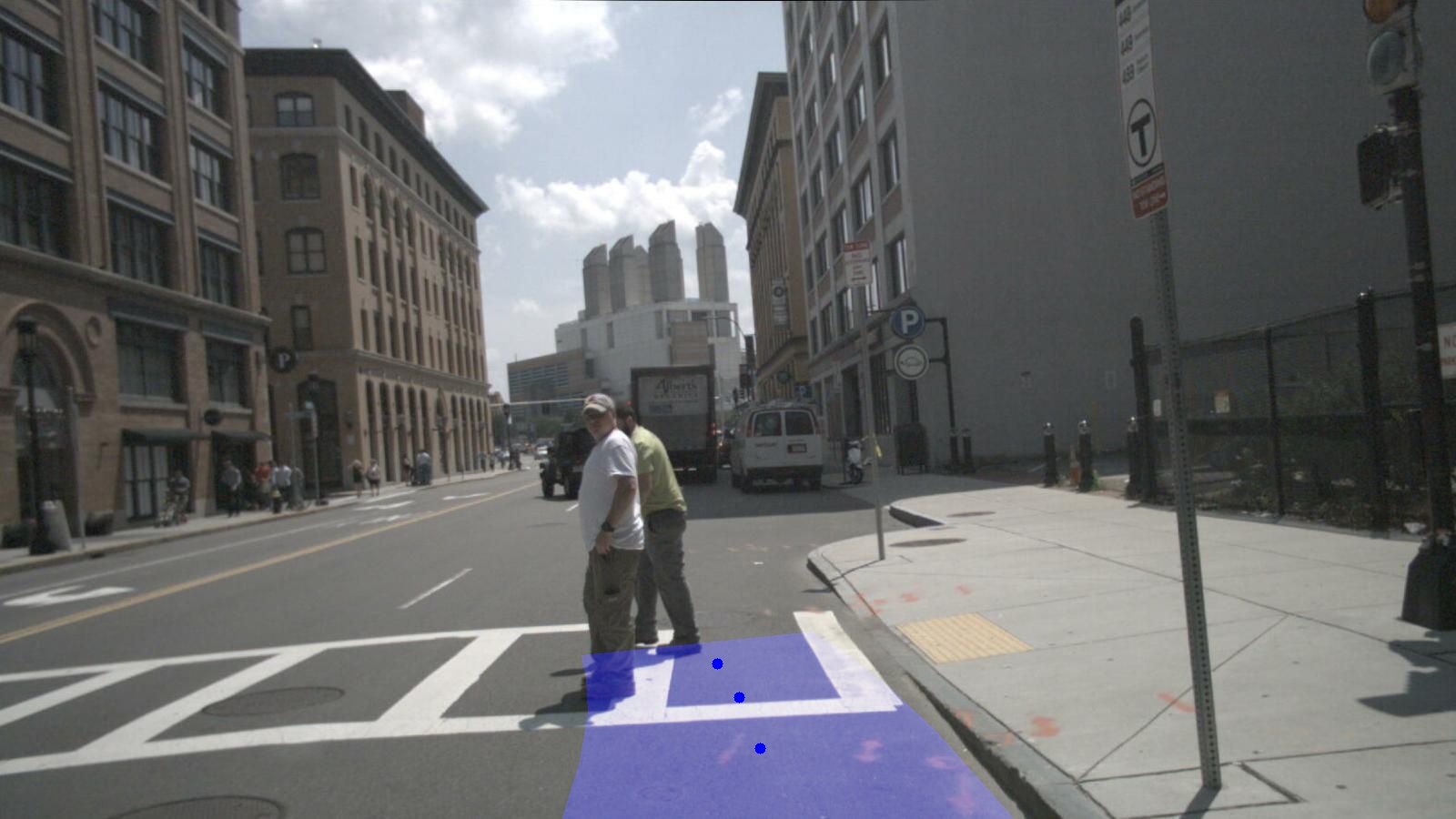} &
        \includegraphics[width=0.425\textwidth]{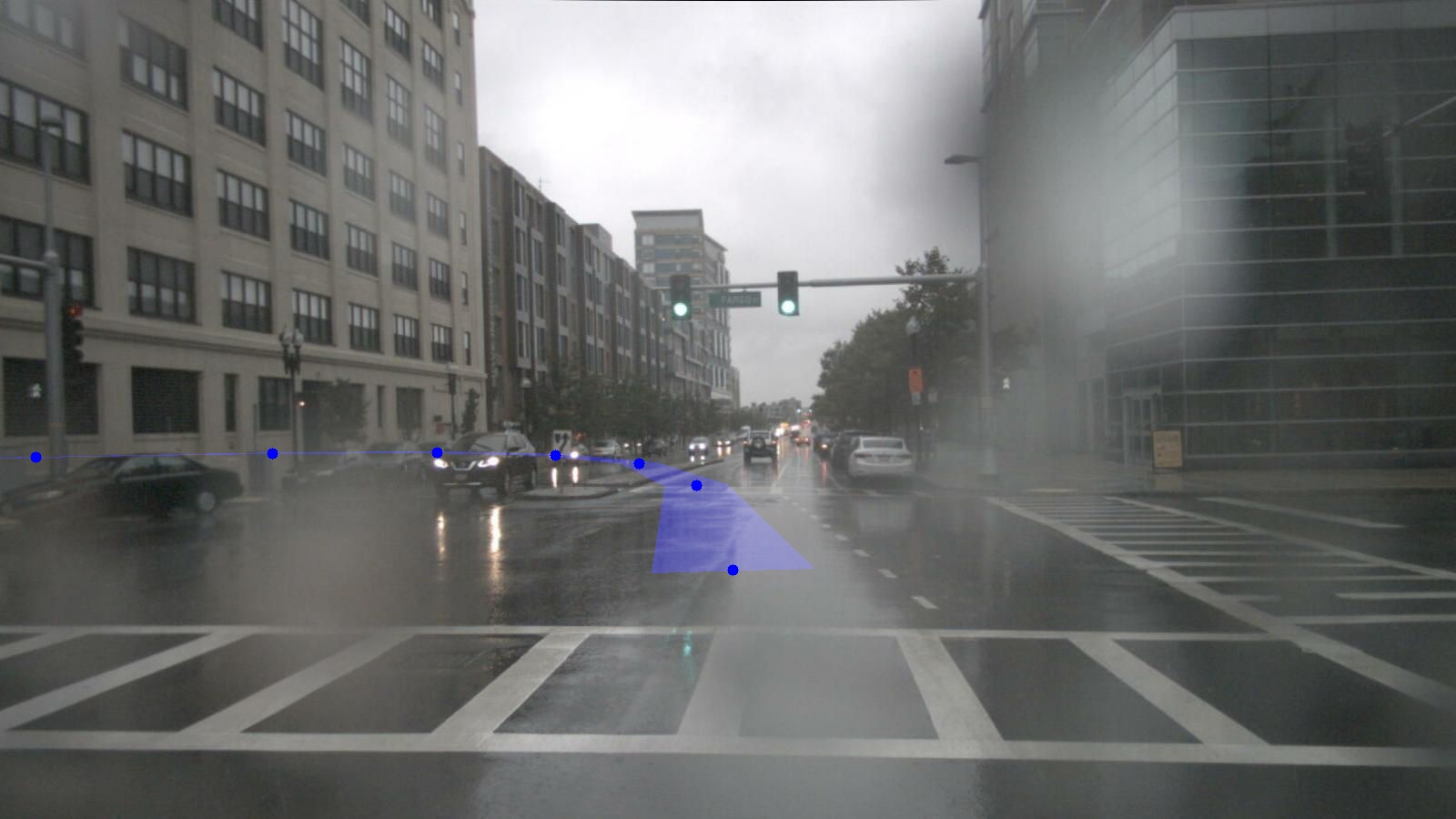} \\
        \textit{No Instruction} &
        \textit{No Instruction} \\[10pt] 
        \includegraphics[width=0.425\textwidth]{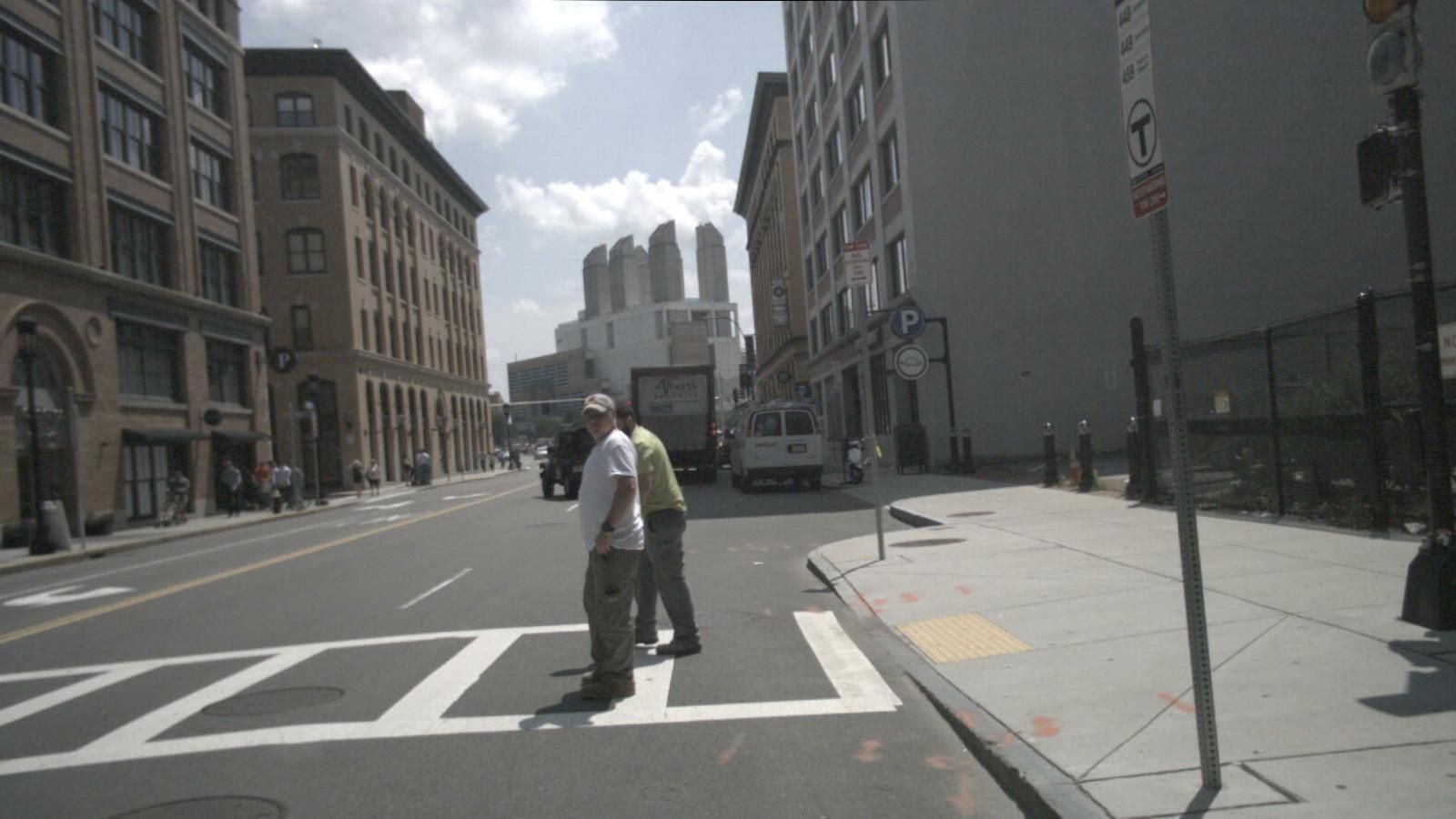} &
        \includegraphics[width=0.425\textwidth]{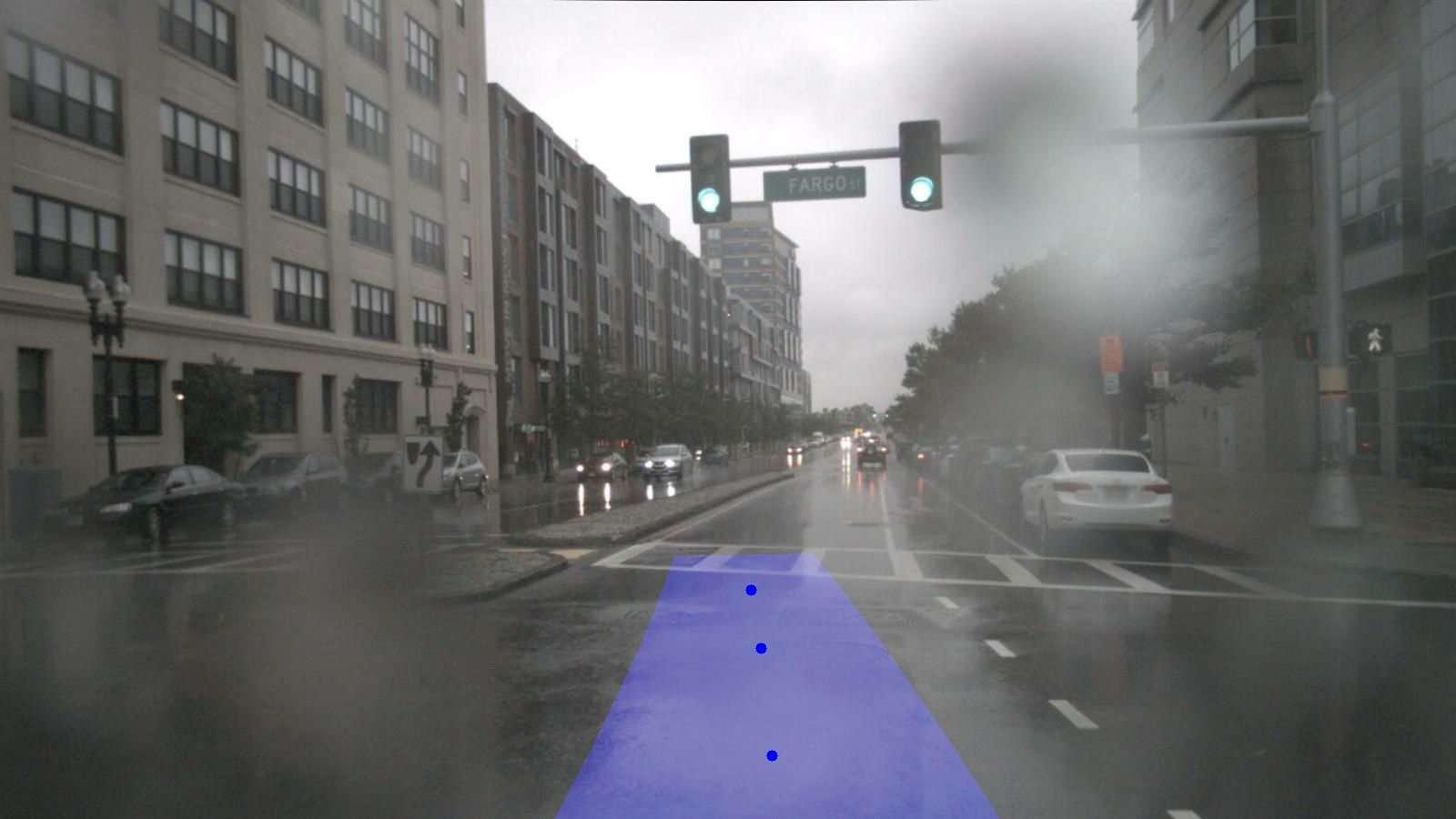} \\
        \parbox{0.425\textwidth}{\centering \textit{``Stop at the curb on the right side of the road right before the crosswalk.''}} &
        \parbox{0.425\textwidth}{\centering \textit{``Go straight when the stoplight turns green''}} \\
        \bottomrule
    \end{tabular}
    \caption{A qualitative comparison of OpenEMMA predictions with and without natural language instruction. Each column shows one scene where the first row is without instruction, and the second row is with its corresponding doScenes prompt. In these cases, the doScenes prompt leads to a much safer plan for the ego vehicle, avoiding passing through an active-use crosswalk or making an unsafe left turn where going straight was expected.}
    \label{fig:qualitative_doscenes}
\end{figure*}

Although the best performing doScenes prompts showed a slight improvement over providing no instruction, certain driving scenarios show a major difference that a well-phrased prompt can have.
Figure~\ref{fig:qualitative_doscenes} illustrates two representative examples where instructions meaningfully corrected OpenEMMA’s behavior.
In Scene 238, the instruction \textit{“Stop at the curb on the right side of the road right before the crosswalk”} allowed the model to understand that the ego vehicle should come to a complete stop and not continue generating waypoints toward the pedestrians crossing the crosswalk. In Scene 642, the directive \textit{“Go straight when the stoplight turns green”} guided the model to proceed generating straight waypoints after the ego car proceed in motion as opposed to incorrectly predicting a left turn into oncoming traffic. These cases show that, despite the small differences in aggregate ADE, natural-language instructions can significantly alter motion behavior in safety-critical contexts, improving both spatial awareness and scene-appropriate planning.


\section{Limitations and Future Research}

doScenes' annotations do not perfectly mirror in-the-moment passenger speech -- many labels are concise and speculative, considering what one \textit{would} have said \textit{a posteriori} rather than what was actually said in the moment (if anything). This introduces a distributional gap between annotated directives and real-world utterances and can reduce instruction compliance in real settings; nonetheless, the annotations still provide a natural language instruction closely aligned to the action observed in the nuScenes data instances.

Our setup treats every doScenes instruction as something the LLaVA model should respond to.
In reality, passenger speech can be unsafe, contradictory, or infeasible \cite{bilius2020multistudy, chen2010application}. Because of this, the ``always-act'' bias can make the model appear more capable than it is, since it still produces a trajectory even when the instruction is unsafe or unrealistic, rather than recognizing that no reasonable action should be taken.

As mentioned in the methodology, our evaluation of doScenes-injected prompt effectiveness relies primarily on ADE. While comparing ADE between the baseline and doScenes-injected runs does provide an initial signal of whether language influences motion, ADE alone cannot confirm that the model truly understood or executed the instruction—it only shows that the resulting path is closer to the path selected by the human driver -- one of many reasonable possibilities or interpretations. For this reason, extension to closed-loop evaluation will be an important future step in evaluating human-responsive autonomous planning. 

As discussed earlier, the effective size of doScenes is substantially reduced once scenes without actionable instructions are removed. This is an inherent limitation that comes with the doScenes dataset, and because of this constraint, we note that our results should be interpreted as a demonstration of feasibility rather than a definitive measure of optimal instruction-conditioned planning performance; as with most data-driven learning tasks, we hypothesize that scaling high-quality training data will scale task performance \cite{ahsan2021effect, subramanian2023towards}.


Our work utilized OpenEMMA with LLaVA-1.6-Mistral-7B as a practical, reproducible starting point. However, newer MLLMs (e.g., Qwen-style VLMs or larger LLaVA variants) may offer stronger reasoning and instruction-following results, and could reveal larger gains from doScenes conditioning.
Additionally, beyond OpenEMMA, it will be important to apply the same instructions to other architectures. Graph-structured models such as DriveMLM, as well as closed-loop systems like LMDrive, would allow for benchmarking whether the patterns we observe in this study hold more broadly rather than being specific results to a single end-to-end model.





\section{Concluding Remarks}

This work provides the first real-world study showing that fine-grained passenger instructions can, measurably, influence an autonomous vehicles motion planning system. By integrating doScenes directives into OpenEMMA, we show that language not only \textit{describes} a scene, but can be used to actively \textit{steer} the trajectory of an autonomous driving system. We find, and describe in our discussion, that differences of trajectory output come from phrasing, clarity, and grounding in the instruction, which reveals a simple but important truth about the value and fragility of language. The autonomous vehicle system takes in natural language from the perspective of the passenger, which makes the model responsive to naturalistic human communication, but also exposes the system to the challenges of interpretation and intent. This research represents a step in connecting expressed human intention and responsive robotic plans; we focus on natural language from inside the cabin as a first challenge of a broader movement made possible through vision-language-action models and novel datasets, namely, exploration of methods for aligning human-in-the-loop directives and motion planning for increased safety and trust in autonomous systems.

\bibliographystyle{IEEEtran}
\bibliography{references}

\end{document}